\documentclass[11pt]{article}

\usepackage[final]{acl}

\usepackage{times}
\usepackage{latexsym}
\usepackage[most]{tcolorbox}

\usepackage[T1]{fontenc}

\usepackage[utf8]{inputenc}
\usepackage{fontawesome5}
\usepackage{hyperref}
\usepackage{soul}
\usepackage{xcolor}

\usepackage{microtype}
\usepackage{enumitem}

\usepackage{inconsolata}

\usepackage{graphicx}

\usepackage{amsmath}
\usepackage{relsize}
\usepackage{ragged2e}

\newcommand{\findmytext}{{\smaller\textsf{FindMyText}}}
\newcommand{\prettyurllarge}[1]{{\normalsize\textsf{\href{#1}{#1}}}}
\newcommand{\prettyurl}[1]{{\smaller\textsf{\url{#1}}}}
\newcommand{\prettyref}[2]{{\smaller\textsf{\href{#1}{#2}}}}
\newcommand{\prettyreflarge}[2]{{\normalsize\textsf{\href{#1}{#2}}}}

\title{\findmytext{}: Robust, Scalable Detection of Text Containment\\ in Large Web-Crawled Corpora}

\author{
  \textbf{Lars Henry Berge Olsen\textsuperscript{1}},
  \textbf{Pierre Lison\textsuperscript{1}},
  \textbf{Martin Jullum\textsuperscript{1}} \and
  \textbf{Mark Anderson\textsuperscript{1}}
\\
  \textsuperscript{1}Norwegian Computing Center, Oslo, Norway. \\ 
  \{\prettyreflarge{mailto:lhbolsen@nr.no}{lhbolsen}, \prettyreflarge{mailto:plison@nr.no}{plison}, \prettyreflarge{mailto:jullum@nr.no}{jullum}, \prettyreflarge{mailto:anderson@nr.no}{anderson}\}@nr.no \\[2mm]
  \faGithub~\prettyurllarge{https://github.com/NorskRegnesentral/FindMyText} \ \ \ \ \ \ \
  \faPlayCircle~\prettyurllarge{https://findmytext.nr.no}  \\[-1mm] \hspace{96mm} {\small (password: EACL1234)}
}

\begin{document}
\maketitle
\begin{abstract}

Identifying whether a text passage is present in a large corpus is central for tasks such as copyright compliance, plagiarism detection, and dataset auditing. This paper presents \findmytext{}, a new method for detecting \textit{text containment} -- that is, for determining whether a text fragment appears, in part or in full, in a given corpus.
The approach builds on prior techniques for document fingerprinting, but extends them with a mechanism to explicitly capture chains of matching fingerprints.  By identifying such chains, the approach can reliably detect near-verbatim copies of a given text. 
Using a benchmarking procedure specifically designed to evaluate text containment, we show that \findmytext{} outperforms alternative methods.
\end{abstract}

\section{Introduction}

Large Language Models (LLMs) depend on colossal amounts of data for their pre-training, most of it gathered through web crawling. Understanding the composition of this data is crucial for tasks such as data selection and curation \cite{Albalak2024-xj, Parmar2024-jo}, enhancing model transparency and interpretability \cite{Wang2023-nk,Chang2025-xo} or detecting copyright infringements \cite{Longpre2025-so,Zhang2026-kw}.

Commercial LLM providers have so far been reluctant to disclose the content of their training data, notably due to legal concerns. This has spurred the development of membership inference attacks on black-box LLMs \cite{Oren2024-id,Shi2024-wi}, often leveraging the models' memorisation abilities \cite{Hartmann2023-ix,Ahmed2026-va}. The reliability of these attacks has, however, been called into question \cite{Meeus2025-yg,Zhang2025-uu,Liu2025-qv}.

\begin{figure}[t!]
\centering
\includegraphics[scale=0.26,trim={2mm 3mm 1mm 0},clip]{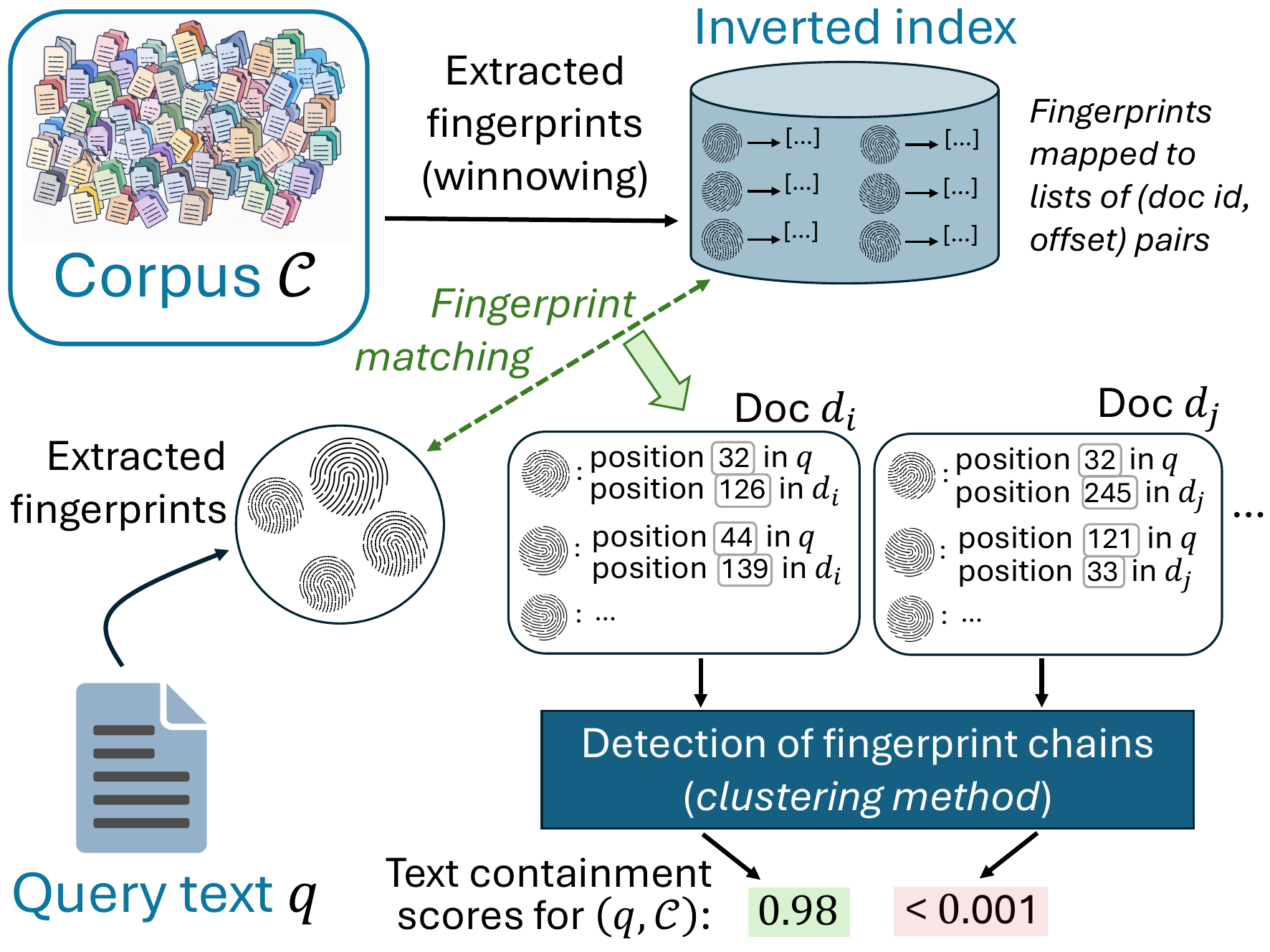}
\caption{General sketch of the \findmytext{} approach. \vspace{-2mm}}
\end{figure}

Transparency obligations introduced by regulations such as the EU AI Act \cite{European-Parliament2024-jf} may in the future compel LLM providers to disclose more details about their data sources. Yet even if the full training data were made publicly available, establishing whether a specific text is part of a large, web-crawled corpus remains technically difficult. A text may be spread across numerous websites, each with its own format, and mixed together with other content. Pre-training data also undergo extensive preprocessing, including text extraction (e.g.~OCR), boilerplate removal, document segmentation, text normalisation and reformatting. Such transformations can defeat standard near-duplicate detection methods.

This paper presents \findmytext{}, a new method for the efficient detection of \textit{text containment}, i.e.~determining whether an input text is wholly or partially contained in a target corpus. This method is specifically designed to (1) scale to large corpora and (2) be robust to discrepancies between the input text and its counterpart in the corpus.  

The approach builds upon \textit{document fingerprinting}, a well-established technique in information retrieval~\cite{Schleimer2003-vi}. However, instead of merely computing the ratio of shared fingerprints, \findmytext{} draws on earlier works in plagiarism detection \cite{Kasprzak2010-pan,Grozea2009-encoplot} and identifies \textit{chains of matching fingerprints}. This enables a more precise detection and localisation of shared fragments, which is essential for tasks such as copyright compliance.

This paper makes the following contributions:
\begin{enumerate}
\itemsep0em 
    \item An efficient, chain-based method for locating shared text fragments (Sec.~\ref{sec:approach}).
    \item A benchmarking procedure to compare text containment methods (Sec.~\ref{sec:benchmark}).
\item An experimental validation of \findmytext{} with this benchmark on three corpora (Sec.~\ref{sec:experiments}).
\end{enumerate}


\section{Related Work}
\label{sec:related_work}

\paragraph{String alignment} Given a query document $q$ and a corpus $\mathcal{C} = \{d_i \text{ for } 1 \leq i \leq |\mathcal{C}|\}$, one straightforward approach to text containment is to use \textit{sequence alignment} to measure the proximity of $q$ with each text $d_i$ composing the corpus $\mathcal{C}$. 

Pairwise alignment algorithms can be divided into \textit{global} and \textit{local} methods. 
Global alignment methods, such as \citet{Needleman1970-ja}, align entire sequences end-to-end. Local alignment, on the other hand, focuses on finding best matching \textit{subsequences}. The best-known local method is the \citet{Smith1981-lg} algorithm. Given two texts $d_1$ and $d_2$ and a scoring scheme, local alignment identifies the pair of regions with the highest alignment score. 

Unfortunately, given their -- typically quadratic -- computational complexity, pairwise alignment algorithms cannot scale to large corpora. Nevertheless, they provide a useful \textit{ground truth} to quantify the degree to which two texts share a subsequence. We notably rely on such a ground truth for the construction of our benchmark (Section \ref{sec:benchmark}). 

\paragraph{Duplicate detection} Information retrieval offers various methods for duplicate detection. If a query $q$ occurs verbatim in a corpus $\mathcal{C}$, exact matching can be performed via hashing. In practice, however, $q$ and its counterpart $d_i$ will often differ slightly, due to e.g.~text normalisation or formatting. 

To handle such variations, \textit{near-duplicate detection} methods typically compare the sets of $k$-grams (or shingles) extracted from the texts. A common measure is the Jaccard similarity, which can be efficiently approximated using compact sketches such as MinHash \cite{Broder1997-or}, which builds a short document signature by retaining only the minimum hash values of its shingles under multiple hash functions. These sketches can then be indexed with locality-sensitive hashing (LSH) to retrieve candidates with high similarity. Alternatively, one can use \textit{winnowing} to select representative fingerprints from the hashes \cite{Schleimer2003-vi}. Unlike MinHash, winnowing preserves positional information, making it well suited for identifying shared passages. Winnowing constitutes the backbone of \findmytext{}, as described in the next section.

\paragraph{Vector-based approaches} Duplicates can also be detected by mapping documents to sparse or dense vector spaces and retrieving texts $d_i \in \mathcal{C}$ close to $q$ under a chosen distance metric. Such representations may rely on compact binary embeddings \cite{Charikar2002-wt}, embeddings tailored for duplicate detection \cite{Zhang2024-np}, or embeddings derived from pre-trained LMs to capture semantic similarity \cite{Abbas2023-wb}. While effective at identifying globally similar documents, these methods are less suited to detect local containment when only small text fragments are shared.

\paragraph{Exact matching at corpus scale} Infini-gram \cite{Liu2024-mu} uses suffix arrays to enable efficient exact $n$-gram retrieval from large corpora. OLMoTrace \cite{Liu2025-olmotrace} extends this approach to trace segments of LM outputs to verbatim matches in training documents. While related to text containment, its goal is to match model outputs to training data rather than assess whether a text is contained within a corpus. Moreover, exact matching may fail when passages contain minor edits, which \findmytext{} is designed to handle.

\paragraph{Plagiarism detection} Plagiarism detection also addresses the retrieval and localisation of reused passages. The PAN shared tasks established a \textit{seed-and-extend} pipeline combining candidate retrieval, passage comparison, and filtering \cite{Potthast2010-pan}. 
\citet{Kasprzak2009-pan} and \citet{Kasprzak2010-pan} retrieve candidate documents with an inverted index of hashed chunks, then locate text reuse regions by aggregating nearby chunk matches. 
\citet{Suchomel2012-pan} extend this work with web search and multi-feature passage comparison. 
ENCOPLOT \cite{Grozea2009-encoplot,Grozea2012-encoplot} represents matches geometrically and grows segments with a Monte Carlo heuristic to find shared passages in candidate document pairs.
Building upon those earlier approaches, \findmytext{} combines a corpus-wide index of winnowed fingerprints with deterministic clustering in positional-offset space, where shared passages appear as chains of fingerprint matches.

\section{Approach}
\label{sec:approach}

\subsection{Fingerprinting}

\paragraph{Winnowing} \findmytext{} first splits a text into words, extracts its shingles (word-level $k$-grams) and converts those into hashes. A text of $n$ words will thus result in $n\!-\!k\!+\!1$ hashes. To reduce the number of hashes, we rely on a \textit{winnowing} mechanism.  Given a hash sequence $h_1,...,h_{n-k+1}$, we slide a window of length $w$ and select, for each window, the minimum hash:
\[
m_i = \min \{ h_j \mid i \leq j \leq i+w-1 \}.
\]
The selected minima are kept as fingerprints, removing duplicates. This greatly reduces the number of hashes to store while ensuring that any shared sequence of length at least $k\!+\!w\!-\!1$ yields a common fingerprint \cite{Schleimer2003-vi}.

\paragraph{Indexing} 
Those fingerprints are then extracted for every document in $\mathcal{C}$, and stored in an \textit{inverted index} $I_{\mathcal{C}}$ that maps each hash $h_i$ to a list of $(doc\_id, pos)$ pairs, where $pos$ is the character offset of the fingerprint in the document. The construction of this index in \findmytext{} can be distributed on many machines, making it possible to scale to large corpora\footnote{In our experiments, indexing a corpus of 171 million documents was done in less than 2 hours on a computing cluster (AMD EPYC), using around 570 CPU-hours.}. Once constructed, this index $I_{\mathcal{C}}$ allows us to efficiently retrieve documents in $\mathcal{C}$ based on fingerprints. 
\subsection{Detection of text containment}
\label{sec:detection}

Let $\mathcal{C}$ be a corpus with indexed winnowed fingerprints and $q$ a query. We aim to determine whether a document $d_i \in \mathcal{C}$ shares text with $q$. To this end, we compute a \textit{text containment score} $s(q,d_i)$ using one of the two strategies described below.

\paragraph{Nb. of shared fingerprints} Let $\mathcal{F}(\cdot)$ represent the set of fingerprints extracted from a document. A simple measure of text containment is the number of unique fingerprints shared by $q$ and $d_i$:
\begin{equation}
s_\text{nb\_shared}(q, d_i) = |\mathcal{F}(q) \cap \mathcal{F}(d_i)| \label{eq:nb_shared}
\end{equation}


The $s_\text{nb\_shared}$ score ignores positional information. Shared segments should, however, contain fingerprints in the same order and (roughly) the same spacing, as captured by the method below.

\begin{figure}[t]
    \centering
    \includegraphics[width=\columnwidth,trim={9mm 6mm 6mm 3mm},clip]{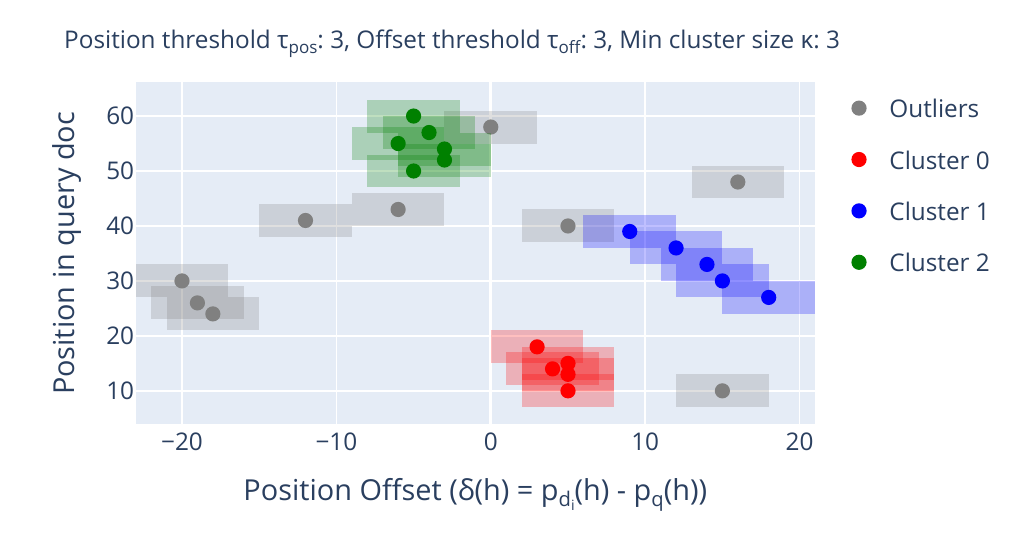} \vspace{-5mm}
    \caption{Illustration of the method for identifying fingerprint chains. Clusters of size $\geq \kappa$ are shown in colour, while fingerprints that cannot be connected to a cluster are ignored. The largest cluster here has 6 fingerprints, yielding a score $s_\text{chain}(q, d_i) = 6$.} 
    \label{fig:clusters}
\end{figure}

\paragraph{Fingerprint chains} Let $h \in \mathcal{F}(q) \cap \mathcal{F}(d_i)$ be a fingerprint found in both $q$ and $d_i$. As winnowing preserves the fingerprint positions, we can retrieve its positions $p_q(h)$ and $p_{d_i}(h)$ in the two texts. Note that a fingerprint may occur more than once in a given text. For each pair of occurrences $\langle p_q(h), p_{d_i}(h)\rangle$ of a shared fingerprint $h$, we define the \emph{position offset} $\delta(h)=p_{d_i}(h)-p_q(h)$. 

Shared fingerprints can then be represented as points $(\delta(h), p_q(h))$ in a two-dimensional space, as illustrated in Fig.~\ref{fig:clusters}. The key intuition is that shared text fragments give rise to \textit{geometrically coherent structures} in this space. More precisely, a text fragment found in both $q$ and $d_i$ will correspond to a sequence of fingerprints with slowly increasing $p_q$ but nearly constant $\delta$.

Formally, we consider two shared fingerprints $h_i$ and $h_j$ as \emph{connected} if $|p_q(h_i) - p_q(h_j)| \leq \tau_\text{pos}$ and $|\delta(h_i) - \delta(h_j)| \leq \tau_\text{off}$, where $\tau_\text{pos}$ controls the maximum gap between two matches, and $\tau_\text{off}$ controls how much the relative shift between the two documents may vary. Connected clusters $\{\mathit{C}_k\}$ are extracted and any cluster with fewer than $\kappa$ fingerprints is discarded. The containment score is then the size of the largest cluster:
\begin{equation}
    s_\text{chain}(q, d_i) = \max_k |\mathit{C}_k| \label{eq:chain} \vspace{-2mm}
\end{equation} 

\paragraph{Normalisation} The raw $s_\text{chain}(q, d_i)$ scores are then normalised. More precisely, we wish to determine whether $q$ and $d_i$ share an aligned passage with a Smith-Waterman score $\geq T$, where $T$ is a predefined threshold.  We first estimate the number of fingerprints $N_{\text{fps}}$ that such a passage is expected to share, based on the following approximations: 
\begin{itemize}
\itemsep0.5em 
    \item Assuming an average word length $\bar{c}$ (around 5.5 for English, if we count whitespace), the number of $k$-grams in a passage of $T$ characters is roughly  $N_{\text{k-grams}} = T/\bar{c} - k +1$.
    \item As the aligned passage may not be fully identical between $q$ and $d_i$, we reduce this count by assuming that a proportion $\epsilon$ of words may change between the two documents, giving a corrected count $N_{\text{k-grams}}' = (1-\epsilon)^k N_{\text{k-grams}}$.
    \item Finally, with a window of size $w$, winnowing keeps on average one $k$-gram every $\frac{w+1}{2}$ positions, thus giving an expected number of shared fingerprints $N_{\text{fps}} = \frac{2}{w+1} N_{\text{k-grams}}'$.
\end{itemize}

Based on those approximations, we model the raw score $s_\text{chain}(q, d_i)$ of a pair $(q, d_i)$ sharing a passage with Smith-Waterman score $T$ as following a normal distribution $\mathcal{N}(N_{\text{fps}}, \phi N_{\text{fps}})$. The dispersion parameter $\phi < 1$ reflects the fact that winnowing selects at least one fingerprint per window, making the count of shared fingerprints more regular than a Poisson variable.

The final containment score, normalised for a given alignment threshold $T$, then becomes:
\begin{equation}
    s_T(q, d_i) = \Phi\!\left(\frac{s_\text{chain}(q, d_i) - N_{\text{fps}}}{\sqrt{\phi N_{\text{fps}}}}\right) \label{eq:pT}
\end{equation}
where $\Phi$ is the standard normal CDF. The word-level edit rate $\epsilon$ and dispersion parameter $\phi$ are fixed empirically.

\paragraph{Corpus-level containment} Finally, to efficiently determine whether at least one document in $\mathcal{C}$ shares a passage with the query text $q$, we:
\begin{enumerate}
\itemsep0em 
    \item Query the index $I_{\mathcal{C}}$ to find the subset of documents $\mathcal{C}_q \subset \mathcal{C}$ for which  $s_\text{nb\_shared}(q, d_i) \geq N_{\min}$, where $N_{\text{min}}$ is a predefined threshold.
    \item Compute $s_T(q, d_i)$ for all $d_i \in \mathcal{C}_q$. The corpus-level containment score is then simply defined as $\max_{d_i \in \mathcal{C}_q} s_T(q, d_i)$. 
\end{enumerate}

\section{Benchmark}
\label{sec:benchmark}

Text containment can be trivially evaluated by sampling documents either from or outside the corpus $\mathcal{C}$. However, this would fail to capture more realistic scenarios, where the query $q$ or its counterpart $d_i \in \mathcal{C}$ have undergone various transformations. To this end, we construct a benchmark of deliberately ``difficult'' instances, either positive or negative:

\paragraph{Positive instances} We first sample a document $d_i$ from $\mathcal{C}$ and select a fragment $q \subset d_i$. The fragment is then subjected to a sequence of editing operations, such as changing word cases or inserting boilerplate\footnote{See Appendix \ref{app:benchmark} for details, together with an example.}. After each edit, a Smith-Waterman local aligner compares $q$ with $d_i$. The process is repeated until the alignment score lies just above a threshold $T_{\text{pos}}$. Finally, the resulting fragment is concatenated with unrelated content. The resulting query $q$ constitutes a positive example, as $q$ shares (by construction) a non-verbatim segment with $d_i \in \mathcal{C}$, although with various edits and unrelated text content. In our experiments, the threshold $T_{\text{pos}}$ was set to $1000$, which corresponds to a text segment of about half a page.  
\paragraph{Negative instances} We again sample a document $d_i$ from $\mathcal{C}$ and initialise $q \leftarrow d_i$. We then iteratively modify $q$ through chunk reordering, LLM-based paraphrasing, or the insertion of boilerplate. The process continues until the alignment score falls below a threshold $T_{\text{neg}}$. The query $q$ is labelled negative, as $q$ and $d_i$ no longer share a segment despite remaining semantically similar. The threshold $T_{\text{neg}}$ was set to $100$, corresponding to the absence of any shared segment longer than 1--2 sentences.

\begin{table*}[t!]
\setlength{\tabcolsep}{4pt} 
    \centering
    \small
    \begin{tabular}{r|rrr|rrr|rrr}
      \textbf{Corpus} $\rightarrow$   & \multicolumn{3}{|c|}{Wikipedia} & \multicolumn{3}{|c|}{arXiv} & \multicolumn{3}{|c}{HPLT} \\
      \textbf{Approach} $\downarrow$& \begin{scriptsize}AUC-ROC\end{scriptsize} & \begin{scriptsize}P@R=90\%\end{scriptsize} & \begin{scriptsize}P@R=99\%\end{scriptsize} & \begin{scriptsize}AUC-ROC\end{scriptsize} & \begin{scriptsize}P@R=90\%\end{scriptsize} & \begin{scriptsize}P@R=99\%\end{scriptsize} & \begin{scriptsize}AUC-ROC\end{scriptsize} & \begin{scriptsize}P@R=90\%\end{scriptsize} & \begin{scriptsize}P@R=99\%\end{scriptsize} \\ \noalign{\vskip 2pt}\hline\noalign{\vskip 2pt}
       Dense Retrieval   & .000 & .474 & .497 & / \ \ & / \ \ & / \ \ & / \ \ & / \ \ & / \ \ \\
       BM25  & .293 & .549 & .561 & .019 & .477 & .500 &  .497 & .501 & .499 \\ 
       MinHash+LSH &  .357 &  .481 & .498  &  .279 & .474 & .498 &  .691  & .536  & .505 \\
       Shared fingerprints  & .748 & .680 & .633 & .364 & .525 & .521 & .831 & .747 & .680  \\
       Chain-based approach  & \textbf{.999} & \textbf{1.000} & \textbf{1.000} & \textbf{.999} & \textbf{1.000} & \textbf{1.000} & \textbf{.999} & \textbf{1.000} & \textbf{1.000} \\
    \end{tabular}
    \caption{Text containments results on the three corpora. 
    For each corpus, a benchmark comprising 1K negative and 1K positive instances was generated (see Sec.~\ref{sec:benchmark}), using $T_{\text{pos}}=1000$ and $T_{\text{neg}}=100$. 
    Results for the dense retriever are only available for the Wikipedia articles due to the availability of \href{https://huggingface.co/datasets/maknee/wikipedia_qwen_4b}{precomputed embeddings} for those texts. As all methods are deterministic, results were obtained with one run on the created benchmark.}
    \label{tab:results}
\end{table*}

\section{Experiments}
\label{sec:experiments}

We evaluate \findmytext{} on three corpora. For each, we construct a benchmark of 1K positive and 1K negative instances as outlined above. 

\paragraph{Corpora} The three corpora are (1) Wikipedia, (2) arXiv papers and (3) generic content crawled from the web. Wikipedia articles are extracted from a \prettyref{https://huggingface.co/datasets/maknee/wikipedia_qwen_4b}{dataset} of 1M English articles crawled in Sept.~2025. The arXiv papers correspond to papers uploaded under CC licenses in Sept.~2024 and included in the Common Pile \cite{Kandpal2026-jn}. Finally, the generic web content is drawn from the English portion of the \prettyref{https://hplt-project.org/datasets/v4.0}{HPLT dataset v4.0} \cite{Burchell2025-dw}. We filter out texts below 2K characters or above 30K characters, yielding 381K articles from Wikipedia, 245K full-text papers from arXiv, and 171M web pages from HPLT.

\paragraph{Metrics} We evaluate each method using \textbf{AUC-ROC}, which measures its ability to distinguish positive from negative examples across all score thresholds, and \textbf{Precision@Recall}, i.e.~the precision achieved at a recall of 0.9 or 0.99.

\subsection{Methods}

\paragraph{MinHash+LSH} 
We use MinHash \cite{Broder1997-or} to estimate the Jaccard similarity between the winnowed fingerprints of the query $q$ and documents in $\mathcal{C}$. For each document, we keep the minimum hash values of the shingles under each of 128 hash functions. 
These are then indexed using locality-sensitive hashing (LSH), which returns candidate documents along with their estimated Jaccard similarities and the top score is recorded for each query. This was implemented with the \begin{small}\textsf{datasketch}\end{small} library \cite{eric_zhu_2024_11462182}.

\paragraph{BM25} We use Okapi BM25 \cite{ROBERTSON1995345} to retrieve documents in $\mathcal{C}$ in which the query terms of $q$ occur often, giving more weight to terms that are rare across the corpus. An instance is labelled positive if the highest BM25 score is above a threshold, and negative otherwise. The implementation relies on the BM25S library \cite{lu2024bm25s}.

\paragraph{Dense retrieval} This baseline represents documents as dense embeddings and retrieves nearest neighbours through approximate vector search \cite{Abbas2023-wb}. We use document embeddings computed with \prettyref{https://huggingface.co/Qwen/Qwen3-Embedding-4B}{Qwen3-Embedding-4B} indexed with \prettyref{https://github.com/facebookresearch/faiss}{FAISS} using HNSW \cite{Douze2025-iq}.


\paragraph{Number of shared fingerprints} This method detects text containment as in Eq. \eqref{eq:nb_shared} -- that is, by counting the fingerprints shared by the query $q$ and a document $d_i$ using the inverted index. 

\paragraph{Chain-based method} The final method is the approach from Eq. \eqref{eq:chain}, which explicitly captures the occurrence of fingerprint chains. The $\tau_{\text{pos}}$ and $\tau_{\text{off}}$ thresholds were both fixed to 30.

\subsection{Results}
\label{sec:results}

Table \ref{tab:results} presents the results. We observe that the two baselines (BM25 and dense retrieval) are misled by the adversarial design of the benchmark. The negative examples were indeed deliberately constructed to be \textit{similar} to one corpus document, but without sharing any text fragments. As those baselines search for documents semantically or lexically similar to $q$, their results can become anti-correlated with the benchmark labels.

The number of shared fingerprints, while performing better than the baselines above, fails as a reliable indicator of text containment, particularly for long texts such as arXiv papers.  As counting shared fingerprints ignores positional information, it cannot distinguish true containment from mere similarity. In contrast, chain-based scores discriminate more effectively between positive and negative examples, achieving a near-perfect AUC. 

The runtime performance of \findmytext{} scales well to larger corpora, as detailed in  Appendix~\ref{app:implementation}.

\section{Conclusion}
\label{sec:conclusion}

We presented \findmytext{}, a method for determining whether a text is partly or fully contained in a large corpus. The method relies on an index of document fingerprints extracted with winnowing. Given a query $q$, it retrieves matching fingerprints and groups them into chains of consecutive matches, providing evidence of shared text fragments. 

The method is designed to be robust to small differences between the query text and the corpus documents.  This is crucial for text containment in web-crawled corpora, which typically undergo extensive cleaning and curation.  

 One central use case for \findmytext{} is to assess whether a web-crawled corpus includes copyrighted or improperly licensed materials. In this context, the goal is to identify reused passages or partial copies, rather than texts that merely convey similar ideas. Text containment is thus closely aligned with the legal notion of \textit{substantial similarity}, which concerns the reproduction of protected expression rather than ideas  \cite{Asay2022-vq}.
 

\newpage

\section*{Limitations}

\findmytext{} aims to facilitate the detection of reused passages and partial copies in large, typically web-crawled corpora. While such evidence may help uncover cases of copyright infringement or plagiarism, it should not be viewed as a comprehensive test for infringement. In particular, \findmytext{} is not intended to detect so-called \textit{derivative works} in which the original content has been extensively rewritten, paraphrased, or translated.

The benchmarking procedure described in Section \ref{sec:benchmark} takes a corpus $\mathcal{C}$ as input and generates a set of positive and negative queries. Positive queries share a substantial text fragment with a corpus document according to the Smith-Waterman aligner, while negative queries do not. The resulting queries are synthetic, although derived from documents in the corpus, and may therefore differ from real-world text-containment queries. However, the benchmark is intentionally constructed to provide ``difficult'' instances: positive queries combine edited corpus fragments with unrelated content, while negative queries remain semantically close to their source documents through chunk reordering and LLM-based paraphrasing. As a result, the benchmark can be viewed as a challenging, near worst-case setting that provides a lower bound on text-containment accuracy.

The experiments were conducted on 3 corpora, each featuring texts of different lengths and styles. One limitation of the evaluation is that the three corpora were all based on English-language documents. Although there is no particular reason to believe the proposed detection method would perform differently in other languages, this generalisation remains to be tested empirically. 

Finally, the results presented in Table \ref{tab:results} were limited to a setting where the two alignment thresholds $T_{\text{pos}}$ and $T_{\text{neg}}$ were respectively set to 1000 and 100. Appendix \ref{app:add_results} provides additional results obtained for other values of $T_{\text{pos}}$.

\section*{Ethical considerations}

\findmytext{} is designed to detect whether a text fragment is contained in a large corpus. The method can support beneficial applications such as copyright compliance, plagiarism detection or dataset auditing. At the same time, text-containment evidence should be interpreted with care. The detection of shared text fragments does not by itself establish copyright infringement, ownership, or legal liability. Such determinations depend on additional legal and contextual factors that are outside the scope of this work. The system should therefore be viewed as a tool for identifying candidate instances of text reuse rather than producing legal conclusions.

\bibliography{pierre_paperpile,biblio}

\newpage

\appendix

\clearpage

\section{Benchmark construction}
\label{app:benchmark}

 We provide here  additional details on the construction of the benchmark introduced in Section~\ref{sec:benchmark}. The key idea is to sample documents from the corpus $\mathcal{C}$, and then iteratively alter the sample through various operations. After each step, we run a local alignment based on the Smith-Waterman algorithm to determine the alignment score between the sample and the original document. Positive instances are obtained by introducing transformations that preserve a substantial text overlap, whereas negative instances are generated by applying transformations until no sufficiently long shared segment remains.

\subsection{Edit operations}

\paragraph{Positive instances.}
A positive query is generated from a consecutive fragment extracted from a document $d_i \in \mathcal{C}$. The fragment is progressively modified and compared to the original document after each transformation. The editing process stops once the alignment score approaches, but does not fall below, a predefined threshold $T_{\text{pos}}$. 

The available transformations are intended to reflect common data curation steps in web-crawled corpora:
\begin{itemize}
\itemsep0em
\item de-hyphenation of words split across line breaks;
\item case normalisation;
\item removal of boilerplate material;
\item standardisation of characters such as quotation marks, ligatures, and dashes.
\end{itemize}

As existing web-crawled corpora only provide us with the "clean" version of the documents, we apply the reverse of those operations to obtain a "noisy" query text $q$ from a corpus document $d_i$.

The transformed fragment is then concatenated with unrelated text content (drawn from outside the documents in $\mathcal{C}$) to form the final query.

\paragraph{Negative instances.}
A negative query is generated by starting from a document $d_i \in \mathcal{C}$ and repeatedly modifying it until its alignment score with the original document falls below $T_{\text{neg}}$. The goal is to remove sufficiently long verbatim passages while keeping much of the original meaning intact.

At each iteration, one of the following operations is applied:
\begin{itemize}
\itemsep0em
\item split the text into chunks (respecting sentence boundaries) and randomly permute their order;
\item replace a selected chunk with an LLM-generated paraphrase;
\item insert short boilerplate passages, such as page headers, footers, or navigation links.
\end{itemize}

Paraphrasing was performed using \prettyref{https://huggingface.co/Qwen/Qwen2-7B-Instruct}{Qwen2-7B-Instruct} with chunks of approximately 400 characters and a sampling temperature of 0.8.

\subsection{Benchmark examples}
\label{sec:benchmark_example}

We provide below two examples of texts in the synthetic benchmark generated for the Wikipedia corpus. 

Example~\hyperref[ex:wiki-positive]{1} is a positive instance in which a text fragment (in green) can be found in both $d_i$ and $q$, although with some discrepancies between the two (underlined in the text) -- for instance, "fi" is transformed into a ﬁ ligature, and some words are hyphenated or have changed their casing. In addition, unrelated content is inserted at the beginning of the text. According to the local aligner, the Smith-Waterman score between $d_i$ and $q$ is 1007, which classifies the pair as a positive example, as its alignment score is above $ T_{\text{pos}}$. 

In contrast, Example~\hyperref[ex:wiki-negative]{2} is a negative instance in which $d_i$ and $q$ do not have a shared text fragment. The query $q$ is created by applying a series of permutations between text chunks, paraphrasing with an LLM, and inserting short boilerplate texts. The query $q$ remains relatively close to the original $d_i$, as it covers the same topic and has many shared words and N-grams. However, the edit operations have led to a document that, according to the local aligner, no longer contains a shared text fragment, with a Smith-Waterman score of only 79 between $q$ and the original $d_i$. 

\clearpage

\begin{tcolorbox}
    [enhanced, width=1.0\textwidth, colback=gray!5, colframe=gray!80, boxrule=0.8pt,
    breakable, title=\textbf{Example 1}: Positive instance in Wikipedia benchmark, before skip=10pt, after skip=10pt]
    \phantomsection\label{ex:wiki-positive}
{\small \textbf{Initial text $d_i$ (from Wikipedia)}:}\\[2mm]
\begin{scriptsize}101 Mitigations" is the fifteenth episode of the thirtieth season of the American animated television series The Simpsons, and the 654th episode overall. The episode was directed by Mark Kirkland with a story by Rob LaZebnik and teleplay by Brian Kelley and Dan Vebber. It aired in the United States on Fox on March 3, 2019. \\[-2mm]
\sethlcolor{green!20}

\textbf{Plot} 

Homer, Bart, Lisa and Maggie are eating at The Gilded Truffle, where they used a misprinted coupon good for 100\% off. Meanwhile, with the saved money, Marge gets a Swedish massage. Outside the restaurant, Homer warns the kids to not hustle to get ahead in life. The valet parking attendant Raphael hands Homer the keys to the wrong car, a fancy 1957 seafoam Cadillac Eldorado Biarritz convertible. They drive away with it, having fun on the road. When they go back to the restaurant, Comic Book Guy accuses them of stealing his father's car, and when he notices his mint condition copy of Radioactive Man \#1 was damaged, he presses charges and Homer is arrested. In court, Judge Snyder finds Homer guilty, even after a touching apology letter written by Lisa. \\[-2mm]

With two weeks before Homer's sentencing hearing, Marge goes to The Android's Dungeon \& Baseball Card Shop to negotiate with Comic Book Guy, who says the issue is about getting respect. \hl{At home, Lisa discovers sentencing mitigation videos, including one employed by Mr. Burns for one of his crimes against Springfield, portraying him in a more sympathetic light as a product of a neglected and bullied childhood. The family's efforts to produce a mitigation video for Homer fall short but Lisa is able to splice together the work using her Final Cut Pro skills. In court, Snyder is initially open to setting Homer free but Comic Book Guy delivers an impassioned courtroom speech, the best Snyder has ever heard. He will deliver his verdict the next day.} \\[-2mm]

\hl{Lisa finds a replacement Radioactive Man \#1 online in a nearby Ogdenville comic book shop, though it fails to quell Comic Book Guy's grudge against Homer. However, he notices Homer's 1975 season one Welcome Back, Kotter keychain, a "precious totem" to Homer, the only gift his father ever gave him. To make Homer feel how he felt with his beloved car and comic, Comic Book Guy smashes the keychain with Thor's hammer Mjölnir, then agrees to drop the charges.} He also befriends Homer and invites him to come to Comic-Con with him. \\[-2mm]

In the epilogue, Bart is shown in detention at school being supervised by Principal Skinner. Bart shows Skinner a prank sentencing mitigation video with Milhouse extolling Bart's transformation, while the video plays Bart escapes from the room. \\[-2mm]

\textbf{Production}

Filmmaker Guillermo del Toro guest starred as himself, directing a video about Mr. Burns. Del Toro previously directed the opening sequence of the twenty-fifth season episode "Treehouse of Horror XXIV." \\[-2mm]

\textbf{Reception}

"101 Mitigations" scored a 0.8 rating with a 4 share and was watched by 2.25 million people, making The Simpsons Fox's highest rated show of the night. \\[-2mm]

Tony Sokol of Den of Geek gave the episode 4 out of 5 stars, stating "What did we learn from this episode? Certainly not the intended lesson that moments of pure joy always have consequences, it is German is earth's closest language to Klingon. The episode is funny and revelatory, though not always tummy rumbling funny. '101 Mitigations' contains a good mix of the clever and the silly, with a moral compass set on cruise control." \\[-2mm]

Dennis Perkins of The A.V. Club gave the episode a C+, stating "For Homer to finally understand the pain his wacky weekly nonsense causes to another person could be a loaded moment, dramatically. But the episode fudges it. The brisk running time—truncated more by del Toro's time-consuming but attention-grabbing cameo—leaves Homer and CBG's rapprochement hanging unsatisfactorily, pawned off on the joke that Homer regards CBG's invitation to Comic-Con as barely preferable to prison. The Simpsons has room for its characters' signature misbehaviors to be deconstructed in a meaningful way. It's a shame '101 Mitigations' doesn't. \\[0mm]

 \end{scriptsize} \hrule \vspace{2mm}
\sethlcolor{green!20}

{\small \textbf{Query text $q$ (insertion of unrelated content + small edits):}} \\[2mm]
\begin{scriptsize}
NACHI Bearing 6908N design bearings have two stamped window-type steel cages, an inner ring without flanges and a guide ring centred on the inner ring. The CC design is indicated by the designation suffix C or CC. Large CC design bearings with the designation suffix EC or ECC have an optimized internal design for increased load carrying capacity. \\[-2mm]

NACHI Bearing 6908N design bearings have two stamped window-type steel cages, an inner ring without flanges and a guide ring centred on the inner ring (d $\leq$ 65 mm) or on the cages (d $>$65 mm). 55 bar rises 30.09 0mph NEE dew-point 54 sunrise 7:09 set 6:58 Autumn New Moon (Blood) More mulch. Two more loads. 3 cubic yards. 4.5 total. This is a lotta mulch, 3 trailer fulls. The good news is the trailer … Continue reading

Monthly Archives: September 2008
60 bar rises 30.07 2mph N dew-point 59 sunrise 7:06 set 7:00 Autumn Waning Crescent of the Harvest Moon rise 5:12 set 6:05 Today and tomorrow will be full gardening days. There are bulbs to plant: daffodils, hyacinths, snow drops, … Continue reading

Plants in position. This is a fruit tree with a guild of plants that will support it. Guilds are a permaculture concept that I will explain later. Paula Westmoreland, a principal in Ecological Gardens, at Day 3 start. More trees … Continue reading

Near end of day 1. Leveling and moving soil. At the end of day 2 we had mounds of soil and plants placed. At the beginning of day 3 Paula Westmoreland gives direction to the crew. Kate cut more carpet … Continue reading

63 bar falls 30.15 0mph NNE dew-point 60 sunrise 7:04 set 7:04 Waning Crescent of the Harvest Moon rise 2:44 set 5:23 What a guy night and day. Last night I relieved Kate after the trailer blew its tire. … Continue reading

Today I have had many opportunities to learn about melding with the movement of the universe. I missed the first lesson when the dumptruck driver put the load of compost in the middle of the truck gate. I could have … Continue reading

65 bar rises 30.02 omph NW dew-point 64 sunrise 7:03 set 7:07 Last Quarter Harvest Moon rise 12:10 set 4:24 Pouring rain. Thunder and lightning. Good for the crops, but if it lasts into tomorrow, not so good for site … Continue reading

66 bar steady 30.00 0mph WNW dew-point 65 sunrise 7:02 set 7:07 Autumn Last Quarter of the Harvest Moon rise 12:10 set 4:34 Orchard schematic from same orientation as photograph below. The large circles are trees, the smaller crenallated … Continue reading

The last day of our site prep. Raking pea gravel, taking up the chain link laid down to protect the yard and moving paving blocks. Tomorrow I plant bulbs. Then, on Wednesday get compost. Thursday and Friday plant and install. … Continue reading \\[-2mm]

62 bar rises 30.11 2mph N dew-point 54 sunrise 6:59 set 7:13 Waning Gibbous Harvest Moon rise 9:58 set 1:37 Starving The Beast By Jennifer Moses Washington Post Tuesday, November 29, 2005; Page A21 BATON ROUGE, La. — “A primary … Continue reading 
\hl{At home, Lisa discovers sentencing mitigation videos, including one employed by Mr. Burns for one of his crimes against Spring}\underline{ﬁ}\hl{eld, }\underline{4/8}\hl{ portraying him in a more sympathetic light as a product of a neglected and bullied childhood. The family's efforts to produce a mitigation video for Homer fall short but Lisa is able }\underline{T}\hl{o splice together the work using her Fi}\underline{-}\hl{
nal Cut }\underline{p}\hl{ro skills. In court, Snyder is initially open to setting Homer free but Comic Book Guy delivers an impassioned courtroom speech, the best Snyder has ever }\underline{9/9}\hl{ heard. He will deliver his verdict the next day. Lisa finds a replacement Radioactive Man \#1 online in a nearby Ogdenville comic book shop, though it fails to quell Comic Book Guy's grudge against Homer. How}\underline{-}\hl{ever, he notices Homer's 1975 season one Welcome Back, Kotter keychain, a "precious totem" to Homer, the only gift his fath}\underline{-}\hl{er ever gave him. To make Homer feel how he felt with his be}\underline{-}\hl{loved car and }\underline{COMIC}\hl{, Comic Book Guy smashes the keychain with Thor's hammer Mjölnir, then agrees to drop the charges.}\\[-2mm]

$\phantom{x}$\end{scriptsize}\end{tcolorbox}
\clearpage

\begin{tcolorbox}
    [enhanced, width=1.0\textwidth, colback=gray!5, colframe=gray!80, boxrule=0.8pt,
    breakable, title=\textbf{Example 2}: Negative instance in Wikipedia benchmark, before skip=10pt, after skip=10pt]
        \phantomsection\label{ex:wiki-negative}

{\small \textbf{Initial text $d_i$ (from Wikipedia)}:}\\[2mm]
\begin{scriptsize}101 Mitigations" is the fifteenth episode of the thirtieth season of the American animated television series The Simpsons, and the 654th episode overall. The episode was directed by Mark Kirkland with a story by Rob LaZebnik and teleplay by Brian Kelley and Dan Vebber. It aired in the United States on Fox on March 3, 2019. \\[-2mm]
\sethlcolor{green!20}

\textbf{Plot} 

Homer, Bart, Lisa and Maggie are eating at The Gilded Truffle, where they used a misprinted coupon good for 100\% off. Meanwhile, with the saved money, Marge gets a Swedish massage. Outside the restaurant, Homer warns the kids to not hustle to get ahead in life. The valet parking attendant Raphael hands Homer the keys to the wrong car, a fancy 1957 seafoam Cadillac Eldorado Biarritz convertible. They drive away with it, having fun on the road. When they go back to the restaurant, Comic Book Guy accuses them of stealing his father's car, and when he notices his mint condition copy of Radioactive Man \#1 was damaged, he presses charges and Homer is arrested. In court, Judge Snyder finds Homer guilty, even after a touching apology letter written by Lisa. \\[-2mm]

With two weeks before Homer's sentencing hearing, Marge goes to The Android's Dungeon \& Baseball Card Shop to negotiate with Comic Book Guy, who says the issue is about getting respect. At home, Lisa discovers sentencing mitigation videos, including one employed by Mr. Burns for one of his crimes against Springfield, portraying him in a more sympathetic light as a product of a neglected and bullied childhood. The family's efforts to produce a mitigation video for Homer fall short but Lisa is able to splice together the work using her Final Cut Pro skills. In court, Snyder is initially open to setting Homer free but Comic Book Guy delivers an impassioned courtroom speech, the best Snyder has ever heard. He will deliver his verdict the next day. \\[-2mm]

Lisa finds a replacement Radioactive Man \#1 online in a nearby Ogdenville comic book shop, though it fails to quell Comic Book Guy's grudge against Homer. However, he notices Homer's 1975 season one Welcome Back, Kotter keychain, a "precious totem" to Homer, the only gift his father ever gave him. To make Homer feel how he felt with his beloved car and comic, Comic Book Guy smashes the keychain with Thor's hammer Mjölnir, then agrees to drop the charges. He also befriends Homer and invites him to come to Comic-Con with him. \\[-2mm]

In the epilogue, Bart is shown in detention at school being supervised by Principal Skinner. Bart shows Skinner a prank sentencing mitigation video with Milhouse extolling Bart's transformation, while the video plays Bart escapes from the room. \\[-2mm]

\textbf{Production}

Filmmaker Guillermo del Toro guest starred as himself, directing a video about Mr. Burns. Del Toro previously directed the opening sequence of the twenty-fifth season episode "Treehouse of Horror XXIV." \\[-2mm]

\textbf{Reception}

"101 Mitigations" scored a 0.8 rating with a 4 share and was watched by 2.25 million people, making The Simpsons Fox's highest rated show of the night. \\[-2mm]

Tony Sokol of Den of Geek gave the episode 4 out of 5 stars, stating "What did we learn from this episode? Certainly not the intended lesson that moments of pure joy always have consequences, it is German is earth's closest language to Klingon. The episode is funny and revelatory, though not always tummy rumbling funny. '101 Mitigations' contains a good mix of the clever and the silly, with a moral compass set on cruise control." \\[-2mm]

Dennis Perkins of The A.V. Club gave the episode a C+, stating "For Homer to finally understand the pain his wacky weekly nonsense causes to another person could be a loaded moment, dramatically. But the episode fudges it. The brisk running time—truncated more by del Toro's time-consuming but attention-grabbing cameo—leaves Homer and CBG's rapprochement hanging unsatisfactorily, pawned off on the joke that Homer regards CBG's invitation to Comic-Con as barely preferable to prison. The Simpsons has room for its characters' signature misbehaviors to be deconstructed in a meaningful way. It's a shame '101 Mitigations' doesn't. \\[0mm]

 \end{scriptsize} \hrule \vspace{2mm}
\sethlcolor{green!20}

{\small \textbf{Query text $q$ (permutations, paraphrasing, small edits):}} \\[2mm]
\begin{scriptsize}
"Number One Hundred And One Countermeasures" serves as the fifteenth installment of the fortieth cycle in the American animated television series, The Simpsons, and stands as the six hundred fifty-fourth episode in total. Nevertheless, he spots Homer's 1975 season one Welcome Back, Kotter trinket, a "cherished emblem" to Homer, the solitary present his parent ever bestowed upon him. Regrettably, "101 Countermeasures" fails to meet expectations. The segment was helmed by Mark Kirkland, who co-authored the narrative with Rob LaZebnik, alongside teleplay contributions from Brian Kelley and Dan Vebber. The launch took place in America on the Fox channel, precisely on March 3, 2019. The storyline progresses chiefly with Homer, Bart, Lisa, and Maggie savoring dishes at The Gilded Truffle, capitalizing on a mistakenly printed voucher for considerable discounts. Simultaneously, Marge employs the leftover money for a rejuvenating Swedish massage. Beyond the eatery, Homer instructs the offspring to avoid rushing to progress in their lives. The valet service operator, Raphael, mistakenly presents Homer with the keys to an incorrect vehicle - an opulent 1957 seafoam-colored Cadillac Eldorado Biarritz convertible. They proceed with it, relishing their journey. The clan's endeavors to create a remediation movie for Homer prove insufficient, yet Lisa succeeds in amalgamating the components utilizing her expertise with Final Cut Pro. In order to evoke Homer's sentiment associated with his cherished automobile and graphic novel, Comic Book Guy employs Thor's mace Mjölnir to destroy the keychain, subsequently consenting to discontinue the allegations. Moreover, he fosters rapport with Homer and proffers an enticement for him to join on an excursion to the comic convention. Thereafter, Bart is depicted in an appendix serving time in school suspension, overseen by the principal, Skinner. Bart delivers a jovial penance video to Skinner, during which Milhouse acknowledges Bart's metamorphosis, yet as it's showcased, Bart engineers an evasion from the vicinity. The exhibition was orchestrated by visiting actor, Luis Guzmán, who portrayed himself, particularly emphasizing Mr. Smith. Guzmán had previously handled the preliminary segment of the television episode, © 2023 Content Authority. All rights are protected. The final episode, "Jungle of Horror XXIV," is highlighted. The "101 Adjustments" critique was given a rating of nil. 8, determined by opinions from four observers, and was observed by precisely two individuals. This production was seen by a grand total of 25 million people, establishing The Simpsons as Fox's most acclaimed show for that specific night. Chris Smith from Empire Magazine rated the show 8 out of 10, pondering "What advantages did this episode offer us?" "Most likely not the supposed lesson that fleeting pleasures invariably entail repercussions, it is German, Earth's nearest language to Esperanto. The sequence is amusing and insightful, yet doesn't consistently provoke laughter. "101 Defensive Tactics" integrates an eclectic blend of humor and inventive elements, underpinned by an unwavering ethical standpoint. Dennis Perkins, a constituent of the trio, representing three out of four members, assessed the segment with a grade of C-, noting "A pivotal moment may emerge when Homer realizes the emotional repercussions of his irregular weekly activities on other characters, narratively speaking. Regrettably, the initiative fails to uphold its assurance. The swift duration, curtailed even more by del Toro's protracted yet captivating guest appearance, results in an unsatisfactory resolution for Homer and CBG's reconciliation, with their truce clumsily disposed of via a jest suggesting Homer finds CBG's Comic-Con invitation marginally more tolerable than incarceration. The animated series, The Simpsons, allows for the dissection of its characters' distinctive rebellious antics in a substantial manner. Within the courtroom, Snyder initially contemplates releasing Homer. However, Comic Book Guy compels the judge with a persuasive speech, the most compelling he has encountered. A decision will be announced the following day. Lisa locates a substitute Radioactive Man \#1 online in a nearby Ogdenville comic store, yet it fails to assuage Comic Book Guy's animosity towards Homer. Upon returning to the eatery, the Comic Book Guy accuses them of appropriating his parent's conveyance. Upon realizing his pristine copy of Radioactive Man \#1 had sustained damage, he initiates legal action, culminating in Homer's incarceration. Recently, within the legal chamber, the document underwent modifications. Judge Snyder emphasizes Homer's accountability, notwithstanding a passionate apology letter dated 8/9 authored by Lisa, disregarding the 5/6 evidence. Approximately Next Page > two weeks ahead of Homer's scheduled court appearance, Marge pays a visit to The Android's Dungeon \& Baseball Card Emporium for conversations with Comic Book Guy, who claims that the fundamental aspect pertains to preserving dignity. © 2018 Enterprise Label. Comprehensive permissions reserved. Upon reaching her dwelling, Lisa discovers documentation for diminution of punishment, inclusive of an artifact employed by individual named Mr. Burns for one of his offenses against Springfield, presenting him in a more empathetic perspective as a consequence of an unattended and tormented youth. Citations Linked resources The Simpson series (30th installment) segments 2019 United States TV content Comic-related Television broadcasts.

$\phantom{x}$\end{scriptsize}\end{tcolorbox}

\newpage

$\phantom{x}$

\begin{table*}[!htbp]
\setlength{\tabcolsep}{4pt} 
    \centering
    \begin{tabular}{r|rrr|rrr|rrr}
      \textbf{Corpus} $\rightarrow$   & \multicolumn{3}{|c|}{Wikipedia} & \multicolumn{3}{|c|}{arXiv} & \multicolumn{3}{|c}{HPLT} \\
      \textbf{Approach} $\downarrow$& \begin{scriptsize}AUC-ROC\end{scriptsize} & \begin{scriptsize}P@R=90\%\end{scriptsize} & \begin{scriptsize}P@R=99\%\end{scriptsize} & \begin{scriptsize}AUC-ROC\end{scriptsize} & \begin{scriptsize}P@R=90\%\end{scriptsize} & \begin{scriptsize}P@R=99\%\end{scriptsize} & \begin{scriptsize}AUC-ROC\end{scriptsize} & \begin{scriptsize}P@R=90\%\end{scriptsize} & \begin{scriptsize}P@R=99\%\end{scriptsize} \\ \noalign{\vskip 2pt}\hline\noalign{\vskip 2pt}
       Shared fingerprints  & .657 & .583 & .500 & .514 & .510 & .500 & .739 & .616 & .500  \\
       Chain-based approach  & \textbf{.980} & \textbf{1.000} & .500 & \textbf{.987} & \textbf{1.000} & .500 & \textbf{.969} & \textbf{1.000} & .500 \\
    \end{tabular}
    \caption{Results on the three corpora when using thresholds on the Smith-Waterman score of respectively $T_{\text{pos}} = 500$ for the positive instances and $T_{\text{neg}} = 100$ for the negative ones to generate the instances of the benchmark. All experimental settings are otherwise identical to Table \ref{tab:results}.}
    \label{tab:results3}
\end{table*}

\begin{table*}[!htbp]
\setlength{\tabcolsep}{4pt} 
    \centering
    \begin{tabular}{r|rrr|rrr|rrr}
      \textbf{Corpus} $\rightarrow$   & \multicolumn{3}{|c|}{Wikipedia} & \multicolumn{3}{|c|}{arXiv} & \multicolumn{3}{|c}{HPLT} \\
      \textbf{Approach} $\downarrow$& \begin{scriptsize}AUC-ROC\end{scriptsize} & \begin{scriptsize}P@R=90\%\end{scriptsize} & \begin{scriptsize}P@R=99\%\end{scriptsize} & \begin{scriptsize}AUC-ROC\end{scriptsize} & \begin{scriptsize}P@R=90\%\end{scriptsize} & \begin{scriptsize}P@R=99\%\end{scriptsize} & \begin{scriptsize}AUC-ROC\end{scriptsize} & \begin{scriptsize}P@R=90\%\end{scriptsize} & \begin{scriptsize}P@R=99\%\end{scriptsize} \\ \noalign{\vskip 2pt}\hline\noalign{\vskip 2pt}
       Shared fingerprints  & .879 & .807 & .778 & .635 & .579 & .553 & .938 & .893 & .843  \\
       Chain-based approach  & \textbf{1.000} & \textbf{1.000} & \textbf{1.000} & \textbf{.999} & \textbf{1.000} & \textbf{1.000} & \textbf{.999} & \textbf{1.000} & \textbf{1.000} \\
    \end{tabular}
    \caption{Results on the three corpora when using thresholds on the Smith-Waterman score of respectively $T_{\text{pos}} = 2000$ for the positive instances and $T_{\text{neg}} = 100$ for the negative ones to generate the instances of the benchmark. All experimental  settings are otherwise identical to Table \ref{tab:results}.}
    \label{tab:results2}
\end{table*}

\newpage

\section{Additional results}
\label{app:add_results}

The results from Table \ref{tab:results} were obtained on the basis of a benchmark where the positive instances were designed so as to have a Smith-Waterman score of at least 1000 between the query document $q$ and a document $d_i \in \mathcal{C}$. This score reflects the presence of a shared text fragment (allowing for some small discrepancies) between the two documents of around half a page. 

We can create alternative benchmarks in which this minimal score for positive instances is modified. Lower scores lead to a more challenging benchmark in which positive instances only share a small common part (and thus a small number of common fingerprints), while higher scores create instances with longer common fragments. 

Tables \ref{tab:results3} and \ref{tab:results2} illustrate the results obtained with a minimum Smith-Waterman score of respectively 500 and 2000. We can observe that the baseline method relying on the number of common fingerprints increasingly struggles when the length of the shared text goes down, especially for long documents such as arXiv papers. Unsurprisingly, longer fragments are easier to detect. 

\newpage 

$\phantom{x}$

\newpage 

\section{Implementation details}
\label{app:implementation}

\begin{small}\textsf{FindMyText}\end{small} is implemented in Python. Its use proceeds in two stages: 
\begin{description}
\itemsep0em 
    \item[Stage 1] Given a corpus $\mathcal{C}$, the user builds an inverted index $\mathcal{I}_{\mathcal{C}}$ of winnowed fingerprints over its documents -- or selects one of the already compiled indexes at their disposal. 
    \item[Stage 2] The resulting index is then queried with a document $q$ to determine whether, and to what extent, its contents can be found in $\mathcal{C}$.
\end{description}

The extraction of winnowed fingerprints is implemented in Numba to maximise efficiency, reaching a throughput per CPU core of approximately 150 ms per million characters\footnote{As measured on an AMD EPYC (2.7-5.7 GHz).}. For our experiments, the size $k$ of $k$-grams was set to 5 (ignoring punctuation), and the window $w$ to 6. 

The construction of the index $\mathcal{I}_{\mathcal{C}}$ can be distributed on many machines, each being responsible for extracting fingerprints on a subset of the full document collection. Those partial results are then merged in a single inverted index in which each fingerprint is mapped to a list of $\left(\textit{doc\_id},\, \textit{pos}\right)$ pairs. To scale to very large corpora, \findmytext{} adopts a disk-based storage scheme built on memory-mapped arrays, allowing it to handle indexes that exceed the available memory while still providing efficient query performance. The lookup for shared fingerprints between $q$ and documents indexed in $\mathcal{I}_{\mathcal{C}}$ is vectorised, thereby retrieving potential matches in a single operation. 

The runtime performance of \findmytext{} is mainly influenced by the index size. The response time was around 16 ms for Wikipedia, 29 ms for arXiv papers, and 373 ms for the much larger HPLT data, which included over 171M texts.\footnote{Runtime measured on an AMD EPYC machine with 64 cores and 160 GB of RAM.} 

Along with the Python package, we provide a web interface for querying the indexes used in our experiments and visualising detected overlaps; see below. 

\newpage

$\phantom{x}$
\newpage

\section{Web interface}
\label{app:web-interface}
\findmytext{} can be tested interactively through a web interface available at \prettyurl{https://findmytext.nr.no/} 
(password: EACL1234).

\begin{figure}[ht!]
\centering
\includegraphics[scale=0.30,trim={0mm 0mm 0mm 0mm},clip]{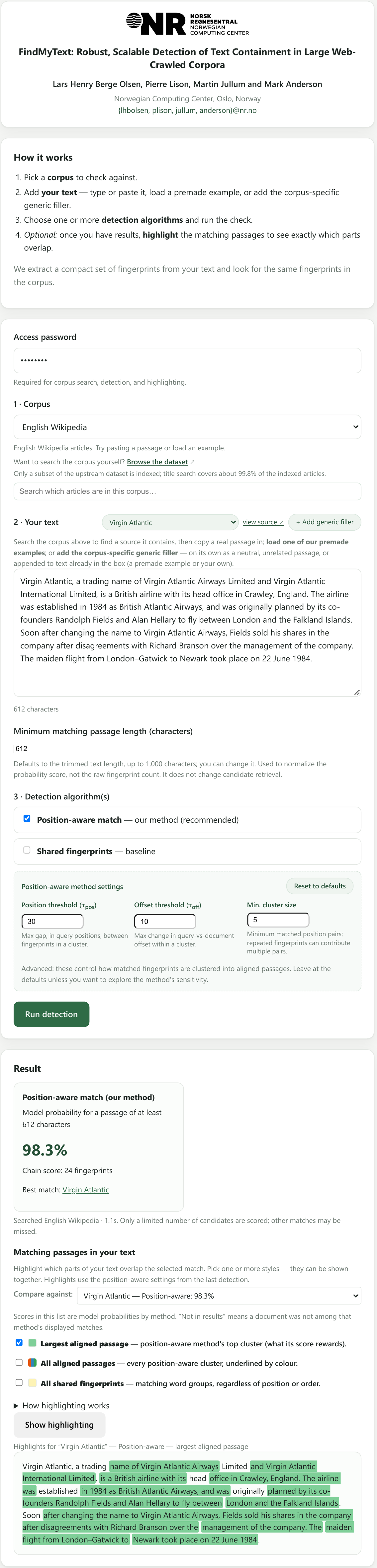}
\caption{FindMyText web interface.}
\label{fig:web-interface}
\end{figure}

Figure~\ref{fig:web-interface} shows a screenshot of this interface, which is made available together with the FindMyText toolkit. The public interface provides precompiled indexes for English Wikipedia and arXiv papers. The HPLT corpus is included in our experiments but is not available in the demo because of the resources required to host its index. The user first selects one of those corpora, and then provides a query text $q$ either by pasting their own text, loading one of the predefined examples, searching the indexed documents for a source text to inspect, or adding corpus-specific filler text. This filler text can be used on its own as a negative example, or appended to a matching passage to create a more challenging query in which the relevant overlap is embedded in unrelated content.

The interface implements the two fingerprint-based methods evaluated in Sec.~\ref{sec:experiments}: the number of shared fingerprints and the chain-based method introduced in Sec.~\ref{sec:detection}. For the chain-based method, the user may also adjust the main clustering parameters, $\tau_{\mathrm{pos}}$, $\tau_{\mathrm{off}}$, and $\kappa$, making it possible to inspect the sensitivity of the clustering step. 
Users can also adjust the minimum passage-length threshold $T$ used in the normalization leading to Eq. \eqref{eq:pT} (default: the trimmed query length, capped at 1,000 characters). Note that this does not affect candidate retrieval or raw fingerprint counts.

After running a query, the interface reports the best matching document for each selected method, together with its score and a short ranking of candidate documents. When metadata is available, the matched documents are shown with human-readable titles and links to the original source or an archived version. The interface can also highlight the matching passages in the query text, either for the largest fingerprint chain, for all detected chains, or for all shared fingerprints. This visualisation makes it possible to inspect whether a high score corresponds to a coherent reused passage, as targeted by the chain-based method, or to more scattered fingerprint overlap.

\newpage

\end{document}